\documentclass[10pt,twocolumn,letterpaper]{article}

\usepackage{cvpr}
\usepackage{times}
\usepackage{epsfig}
\usepackage{graphicx}
\usepackage{amsmath}
\usepackage{amssymb}

\usepackage[pagebackref=true,breaklinks=true,letterpaper=true,colorlinks,bookmarks=false]{hyperref}

\cvprfinalcopy 
\def\cvprPaperID{****} 

\begin{document}

\title{Real-time Face Mask Detection in Video Data}

\author{
Yuchen Ding, Zichen Li, David Yastremsky\\
University of Pennsylvania\\
{\tt\small \{ycding\} \{zachli\} \{dyas\} @seas.upenn.edu}
}

\maketitle

\begin{abstract}
In response to the ongoing COVID-19 pandemic, we present a robust deep learning pipeline that is capable of identifying correct and incorrect mask-wearing from real-time video streams. To accomplish this goal, we devised two separate approaches and evaluated their performance and run-time efficiency. The first approach leverages a pre-trained face detector in combination with a mask-wearing image classifier trained on a large-scale synthetic dataset. The second approach utilizes a state-of-the-art object detection network to perform localization and classification of faces in one shot, fine-tuned on a small set of labeled real-world images. The first pipeline achieved a test accuracy of 99.97\% on the synthetic dataset and maintained 6 FPS running on video data. The second pipeline achieved a mAP(0.5) of 89\% on real-world images while sustaining 52 FPS on video data. We have concluded that if a larger dataset with bounding-box labels can be curated, this task is best suited using object detection architectures such as YOLO and SSD due to their superior inference speed and satisfactory performance on key evaluation metrics. 
\end{abstract}

\section{Introduction}

In 2020, the largest pandemic in recent history spread through the world: COVID-19. As of May 1st, 2021, there have already been 152 million cases and 3 million deaths around the world \cite{noauthor_covid-19_nodate}. In many regions, those numbers are considerably under-counted \cite{charlie_excess_2021}. Beyond that, many parts of the world have slowed or stopped due to the human, economic, and social impacts of distancing and protection measures. For the purpose of the ongoing pandemic and predictions for future pandemics \cite{dodds_disease_2019}, this project seeks to create a mask detection system that is capable of recognizing whether people in surveillance-type video streams are correctly wearing their masks.

\subsection{Pipeline Overview}

Due to the real-time and real-world deployment constraints of such task, we decided to tackle this problem from two fronts - performance and efficiency. The first pipeline that focuses on accuracy uses a pre-trained face detector to extract faces from the frame, then passes the cropped faces to an image classifier. This mask-wearing classifier is trained on a large-scale synthetic dataset of 180,000 images divided into three classes: mask correctly worn, mask incorrectly worn, and no mask worn. We experimented with various models for this classifier, from traditional machine learning approaches such as random forest and Haar-cascades to state-of-the-art computer vision architectures such as DenseNet and ResNet.

The second pipeline leverages a real-time object detection architecture called YOLO, which is short for ``You Only Look Once''. As implied by the name, this is an extremely fast and efficient model designed specifically for object localization and classification in real-time settings. We trained this model on a real-world dataset with face bounding box labels and mask-wearing classifications. Due to the time-consuming process of bounding box annotation and the lack of methods to generate synthetic images, this dataset is comparatively small with 14,233 images.

\subsection{Related Works}
Projects with similar intent have been quite popular due to the ongoing pandemic. In the paper by Adnane Cabani and his colleagues from Universite de Haute-Alsace, a method was proposed to utilize haar-cascade based feature detectors to individually determine the presence of nose and and mouth from a detected face \cite{cabani_maskedface-net_2021}. Their logic follows that no mask is worn if we can successfully detect a mouth from the face, mask is worn incorrectly if we can detect a nose by not a mouth, and mask is worn correctly if we can detect neither a nose nor a mouth. This approach is efficient and intuitive but has severe limitations - it can only process full-frontal faces and one can easily trick the detector by covering their mouth and nose with their hand. 

Another approach is proposed by Chandrika Deb in his Github project \cite{deb_facemaskdetection_2021}. Similar to our first proposed pipeline, he utilizes a Caffe-based face detector in conjunction with a fine-tuned MobileNetV2 for mask-wearing classification. He was able to achieve a decent 0.93 f1-score on the classifier. Nevertheless, he used a very small dataset of 4095 images, which might not be representative of different ethnicities, genders, and types of facial coverings that the system might encounter in real-world settings. His data was also split into only two classes: with mask and without mask. So his model is incapable of detecting if someone is incorrectly wearing their mask (i.e having his or her mask below the nose). 

Lastly, the Github project by the Chinese company AIZOO Tech uses a object detection network for both face mask detection and classification, similar to our second proposed pipeline \cite{aizootech_facemaskdetection_2021}. They used a lite version of SSD, which is short for Single-Shot Multi-box Detector \cite{liu_ssd_2016}, and achieved around 0.9 in precision and recall. This architecture is very similar to YOLO in both their underlying principal and their intended applications. However, the original SSD was published in 2015, which could be very outdated compare to the fifth iteration of YOLO that we are using, which is published in late 2020. Additionally, their dataset contains only 4095 labeled images - which could be too small for the reasons discussed previously.  

\section{Face Detector $\rightarrow$ Classifier Pipeline}
We need two networks for this pipeline - a face detector and a mask-wearing image classifier. Since face recognition has already been a well-defined and established task in computer vision with many existing solutions - there is no need to reinvent the wheel. So for the face detector, we turn to an existing face recognition package built on-top of PyTorch called Facenet \cite{esler_facenet-pytorch_2021}. This package contains multiple pre-trained deep learning face detectors and we will specifically be using MTCNN, which is short for Multi-task Cascaded Convolutional Networks, due to its superior speed and efficiency \cite{zhang_joint_2016}. As for the image classifier component that our project mainly focuses on, we will test various models from traditional machine learning algorithms to state-of-the-art deep learning architectures. The experiment process and results will be explained in the next few sections.

\subsection{Dataset and Pre-processing}

For the image classification pipeline, we used the MaskedFace-Net dataset created by Université de Haute-Alsace, which contains 133,784 artificially generated images of size 1024 by 2014, each containing a single human face that are either correctly wearing a mask or incorrectly wearing a mask \cite{cabani_maskedface-net_github_2021}. These images have been artificially generated by placing a blue medical mask over images of uncovered frontal faces. To complete the third category where no mask is worn, we used 70,000 images from the Flickr Face Dataset develoepd by NVIDIA AI Lab \cite{tero_nvlabsffhq_2021}. So in summary, there are 67,049 instances in the correctly masked category, 66,734 instances in the incorrectly masked category, and 70,000 instances in the no mask category.

The first step in pre-processing was to resize all images from 1024x1024 to 128x128, in order to scale down the original 38GB input. After the data was reduced to a more manageable size, we randomly split the whole dataset into train, validation, and test set with a ratio of 80:10:10. Lastly, we performed data augmentation for translational and scale in-variance. The training images were first normalized then passed through a series of built-in PyTorch transformations including ColorJitter, Random Rotation, RandomResizedCrop, GaussianBlur, and RandomErasing. This was done to minimize the overfitting of our model to training data. 

\subsection{Classifier Baseline}
Before looking at advanced neural network architectures, we wanted to see whether simpler models would achieve high performance. Given that models tend to have a trade-off between performance and speed, starting with simpler models allowed us to get a baseline for performance (see Table 1). Haar-based facial feature detection was one approach we tried, as this is a classic tried-and-true pattern recognition algorithm for detecting facial features \cite{viola_rapid_2001}. We also applied a random forest with a max depth of 2. As an ensemble method, random forests has great performance and prevents overfitting, while also having the advantage of being highly efficient. Given that mask wearing would be dependent on the pixels at the center of the image, it is possible that random forests would learn these feature combinations. And it appears they did - the random forest model achieved 94.33\% accuracy, performing worst for predicting incorrectly masked folks at 87\% and best for predicting correctly masked folks at 99\% accuracy. It was also by far our fastest model, performing 6951 instances per second on CPU. Our last base model was a vanilla CNN with two convolutional layers, each followed by a leaky ReLU activation function and a max pooling layer. These were followed by two linear layers, also using leaky ReLU as the activation function. This model achieved 98.55\% test accuracy.

\begin{table*}
\begin{center}
\begin{tabular}{|l|c|c|c|}
\hline
 Model & Test Accuracy (class-wise) & Test Accuracy (total) & Inferences/Sec \\
\hline\hline
Haar-cascade & .90/.49/.79 & 0.7266 & 45.95 (CPU) \\ 
Random Forest & .99/.87/.97 & 0.9433 & 6951.18 (CPU) \\
Vanilla CNN & N/A & 0.9855 & 775.35 (V100) \\
\hline
\end{tabular}
\end{center}
\caption{Baseline Model Performance}
\end{table*}

\subsection{Advanced Models}
In determining the best state-of-the-art deep learning architectures to use for transfer learning, we wanted to weigh accuracy versus network size. As a proxy for size, we consulted graphs that plotted accuracy against the number of operations and number of parameters (see Figure 1). We looked for models in the upper-left quadrant with few parameters, as indicated by the bubble size. As a result, we picked DenseNet161 \cite{huang_densely_2018}, MobileNet v2 \cite{howard_mobilenets_2017}, Inception v3 \cite{szegedy_rethinking_2015}, and ResNet18 \cite{he_deep_2015}. 

\begin{figure}[t]
\begin{center}
   \includegraphics[width=1.0\linewidth]{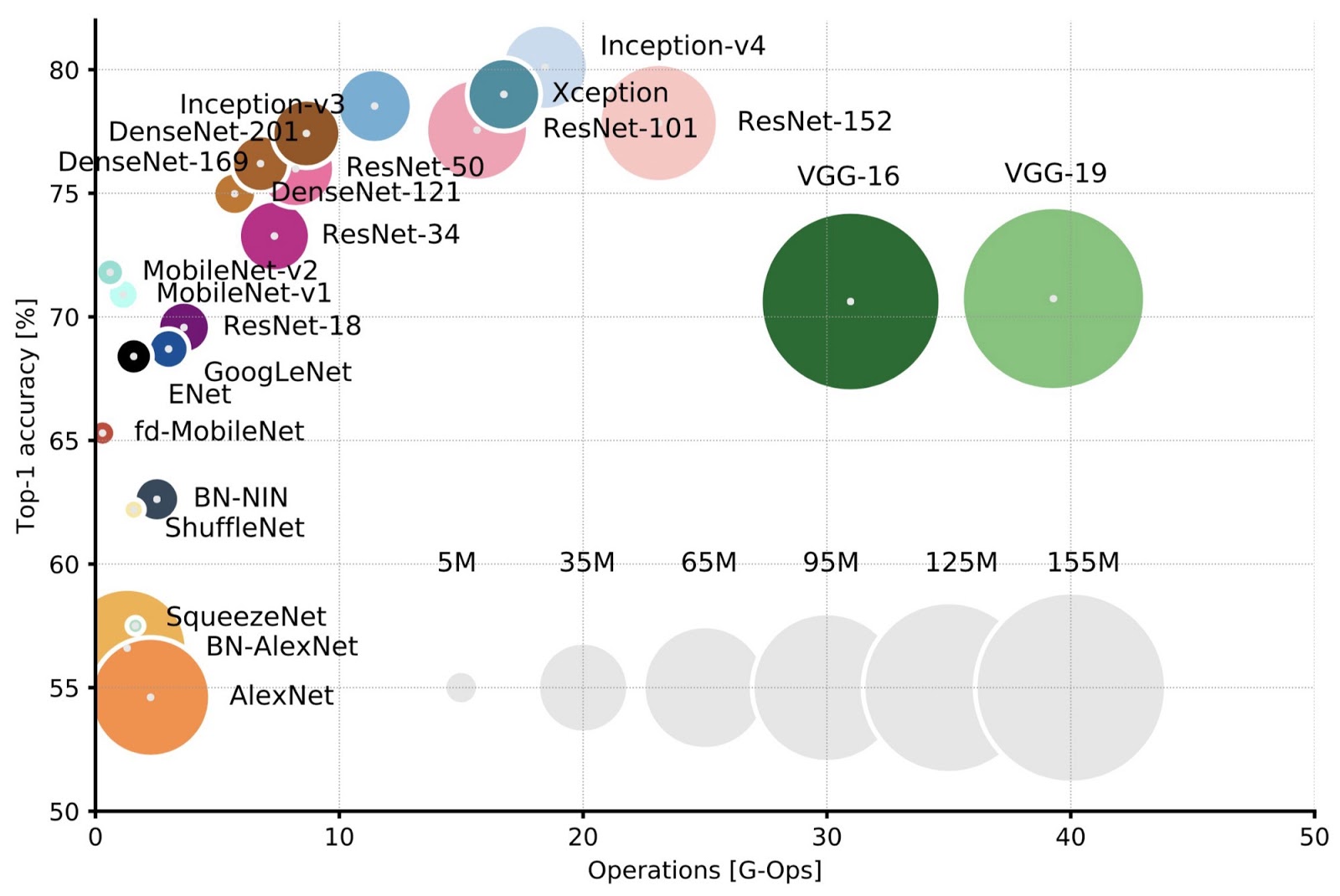}
\end{center}
   \caption{Top-1 one-crop accuracy versus amount of operations required for a single forward pass in multiple popular neural network architectures \cite{culurciello_analysis_2018}.}
\label{fig:long}
\label{fig:onecol}
\end{figure}

\begin{table*}
\begin{center}
\begin{tabular}{|l|c|c|}
\hline
 Model & Test Accuracy (total) & Inferences/Sec (V100) \\
\hline\hline
ResNet18 & 0.9975 & \textbf{680.83} \\ 
MobileNet v2 & 0.9995 & 577.15 \\
DenseNet161 & \textbf{0.9997} & 301.49 \\
Inception v3 & 0.9983 & 425.99 \\ \hline
CNN (Distillation) & 0.9985 & 775.35 \\
\hline
\end{tabular}
\end{center}
\caption{Advanced Model Performance}
\end{table*}

Here we will give a brief architecture highlight of the four networks that we chose. ResNet18 makes use of shortcut connections that are built upon residual blocks, containing about 11 million parameters. MobileNet v2 is a lightweight model that only has 3 million parameters. It uses depth-wise separable convolution and linear bottlenecks between the layers. For DenseNet, each layer receives feature maps from all preceding layers so that feature propagation is strengthened. The total number of parameters in this architecture is 29 million. InceptionNet v3 is a much wider network architecture that has about 24 million parameters. Multiple kernels of small and different sizes are implemented within the same layer.

As shown in Table 2 summarizing the performance of the advanced models, the test accuracies across the board are significantly higher than the base models, and they are all above the 99.5\% accuracy line. The highest accuracy of 99.97\% was achieved by DenseNet, and the second highest accuracy of 99.95\% was obtained by MobileNet v2, followed by the 99.83\% accuracy of InceptionNet v3, and the lowest accuracy of 99.75\% was by ResNet18. While looking at the inference speed of these models, ResNet was the fastest, which was able to inference 680 instances per second on Tesla V100, and DenseNet was the slowest mode, which was capable of inferring only 301 instances per second. The trade-off between accuracy and speed was noticeable here.

\subsection{Distillation}
As our use case required model speed as well as performance, we used the technique of knowledge distillation to see if we could achieve comparable performance with a smaller model. Using our initial CNN model as the student, we performed vanilla distillation using our highest-performing model, DenseNet (99.97\% test accuracy), as the teacher network. The result was a CNN with 99.85\% test accuracy and the best inference speed out of all our advanced models - 775.35 instances per second on a Tesla V100. The CNN had 15\% as many parameters as DenseNet. Given the small size, fast computation, and high accuracy, the CNN after distillation met our project goals and would be a good model to use going forward.

\section{Object Detection Pipeline}
\subsection{Curating Dataset}
To train or fine-tune an object detection network like SSD and YOLO, we need a dataset with ground-truth bounding box labels in addition to the mask-wearing classifications. This makes gathering a large dataset very challenging as the annotation process is time-consuming and there are no straight-forward methods to generate synthetic data like the first pipeline. Luckily we were able to combine two bounding-box-labeled datasets on Kaggle - one from Wobot Intelligence \cite{wobot_face_2020} and another from Andrew Moranhao \cite{maranhao_face_2017}. However we noticed that the ``no mask worn'' class is severely under-represented in both of the datasets. To address this, we downloaded about 3000 images from the COCO dataset \cite{lin_microsoft_2015} that contains the label ``Person'' and ran our MTCNN face detector discussed previously to artificially generate facial bounding boxes. We were careful in only keeping detection outputs with a confidence score above 0.9 - so this method should be robust enough to generate pseudo ground-truth labels for the ``no mask worn'' class. At the end, our custom dataset contains 7364 positive instances (mask correctly worn) and 6869 negative instances (mask incorrectly worn + no mask worn).
\subsection{YOLO v5}
For the object detection architecture, we decided to use the fifth iteration of YOLO, which is short for You Only Look Once \cite{jocher_ultralyticsyolov5_2021}. The original paper by Redmon et. al was published in May 2016, and they were the first object detection network to combine the problem of object localization (bounding boxes) and classification in one end-to-end differentiable network \cite{redmon_you_2016}. The underlying principle is that YOLO treats detection as a regression problem - it divides the image into a grid, and for each grid cell it simultaneously produce boxing box confidence and class probability (see Figure 2). The model then aggregates those results to produce the final bounding box output and classification. This architecture is known for its performance and efficiency on real-time video data, so it is naturally a good fit for our task. 

\begin{figure}[t]
\begin{center}
   \includegraphics[width=1.0\linewidth]{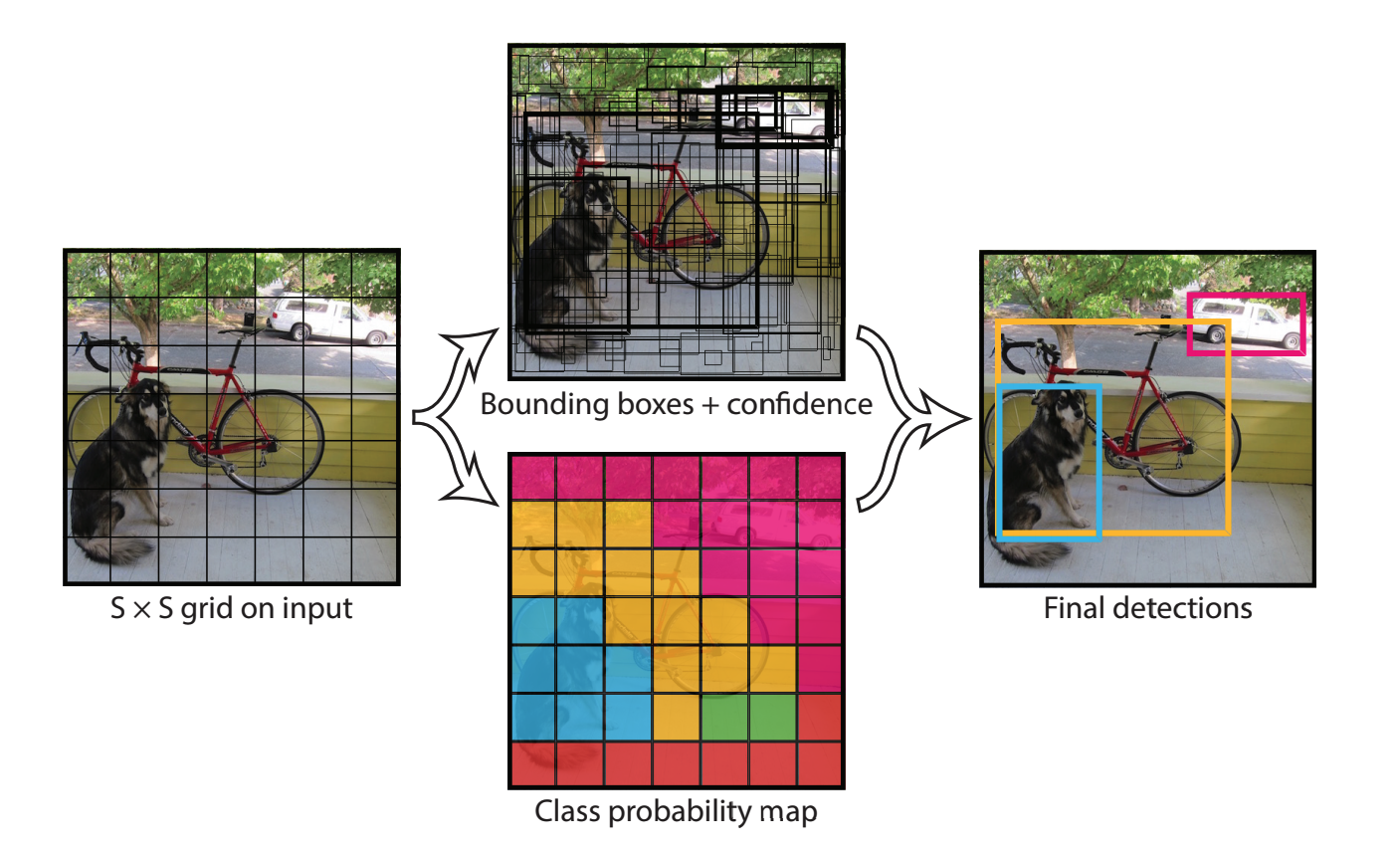}
\end{center}
   \caption{YOLO Algorithm Overview \cite{redmon_you_2016}}
\label{fig:long}
\label{fig:onecol}
\end{figure}

Over the years, there have been multiple iterations and improvements over the original YOLO architectures. This leads us to the fifth version, YOLO v5, developed by the company Ultralytics. This latest iteration utilizes Cross Stage Partial Network (CSPNet) \cite{wang_cspnet_2019} as the model backbone and Path Aggregation Network (PANet) \cite{liu_path_2018} as the neck for feature aggregation (see Figure 3). These improvements have led to better feature extraction and a significant boost in the mean averaged precision score.

\begin{figure}[t]
\begin{center}
   \includegraphics[width=1.0\linewidth]{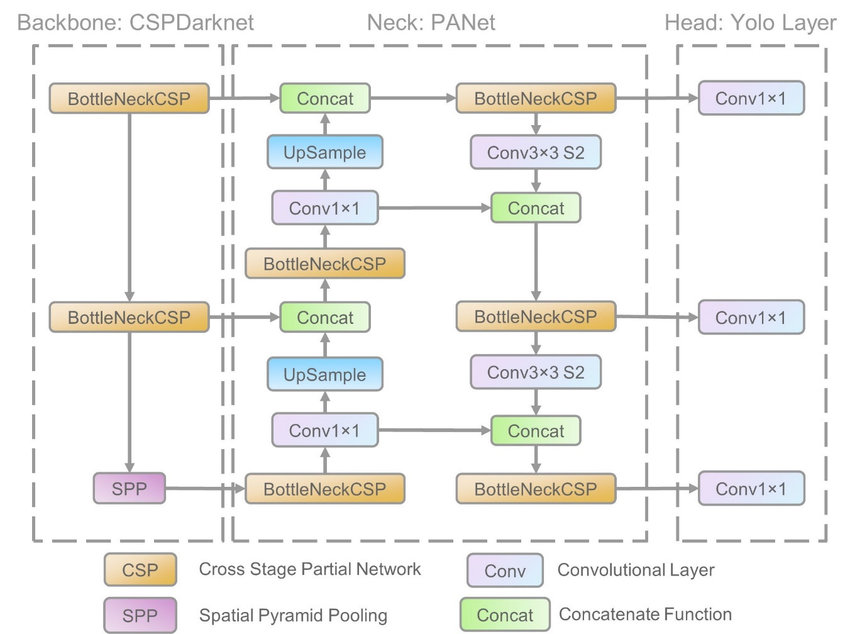}
\end{center}
   \caption{YOLO v5 Architecture Overview \cite{noauthor_overview_nodate}}
\label{fig:long}
\label{fig:onecol}
\end{figure}

\subsection{Experiment Results}
From Table 3 showing the performance evaluation of our fine-tuned YOLO v5 model, we can see that it does very well on the task of face mask detection by achieving a mAP score of 0.898 across both classes on the test data. Looking at the validation loss curves, we can see that it is also learning both the bounding box prediction task and the mask-wearing classification task. 

\begin{figure*}
\begin{center}
\includegraphics[width=1.0\linewidth]{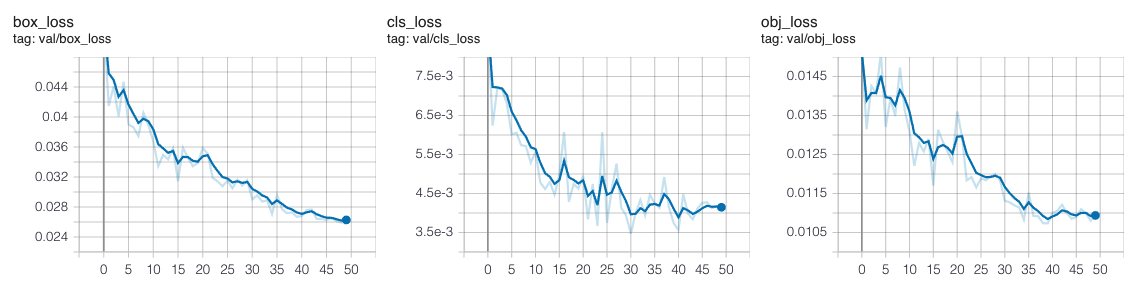}
\end{center}
   \caption{YOLOv5 Validation Loss Curves}
\label{fig:short}
\end{figure*}

\begin{table*}
\begin{center}
\begin{tabular}{|l|c|c|c|}
\hline
 & Negative mAP & Positive mAP & Total mAP \\
\hline
Train & 0.974& 0.964 & 0.969 \\ 
Validation & 0.869 & 0.908 & 0.888 \\
Test & 0.894 & 0.902 & 0.898 \\
\hline
\end{tabular}
\end{center}
\caption{YOLOv5 Mean Averaged Precision on Test Data}
\end{table*}

\section{Analysis on Real-World Data}
Now that we have both of our face mask detection pipelines that were trained on different datasets, how do they compare when run on real-world video data? To test this we gathered two videos - one is a person sitting in front of a webcam taking on and off his mask, and the other is a person filming his walk along the Brooklyn Bridge during the pandemic \cite{youtube_brooklyn_2021}. The webcam video serves as a baseline in a very-controlled environment with only a single face close to the camera. The Brooklyn Bridge video resembles more closely to a setting where the face detection system could potentially be deployed (i.e in public surveillance). 

We ran both pipelines on these two videos, their sample outputs are shown in Figure 5 and Figure 6. YOLOv5 maintained an incredible 52 FPS on both sets of video data while MTCNN + ResNet18 of the first pipeline only achieved 6 FPS. Visually verifying the prediction accuracy, we observed that both pipelines performed almost flawlessly on the webcam video. YOLOv5 had a more stable detection output throughout the whole video. In the Brooklyn Bridge video, we see YOLOv5 completely outperform MTCNN + ResNet18. YOLOv5 was able to accurately detect and classify all face instances that appeared in the video. MTCNN + ResNet18, on the other hand, not only had numerous misdetection of faces but also only produced the correct classification when the person is wearing blue medical mask and extremely close to the camera.

Upon closer examination, we observed that a traditional face detector like MTCNN might fail to detect a face when the person has a mask and other types of facial coverings on such as sunglasses or a hat. These coverings occludes key facial features thus making the pipeline fail before it can even reach our mask-wearing image classifier. In contrast, YOLOv5 was designed specifically to overcome this challenge. It was \textbf{not} trained to first detect faces then classify if the face has a mask on. Instead it has learned to detect directly ``face with correctly worn mask'' and ``face with incorrectly worn mask''. This key difference enables single-shot architectures like YOLO to excel in the tasks of object localization and recognition. 

\begin{figure*}
\begin{center}
\includegraphics[width=0.33\linewidth]{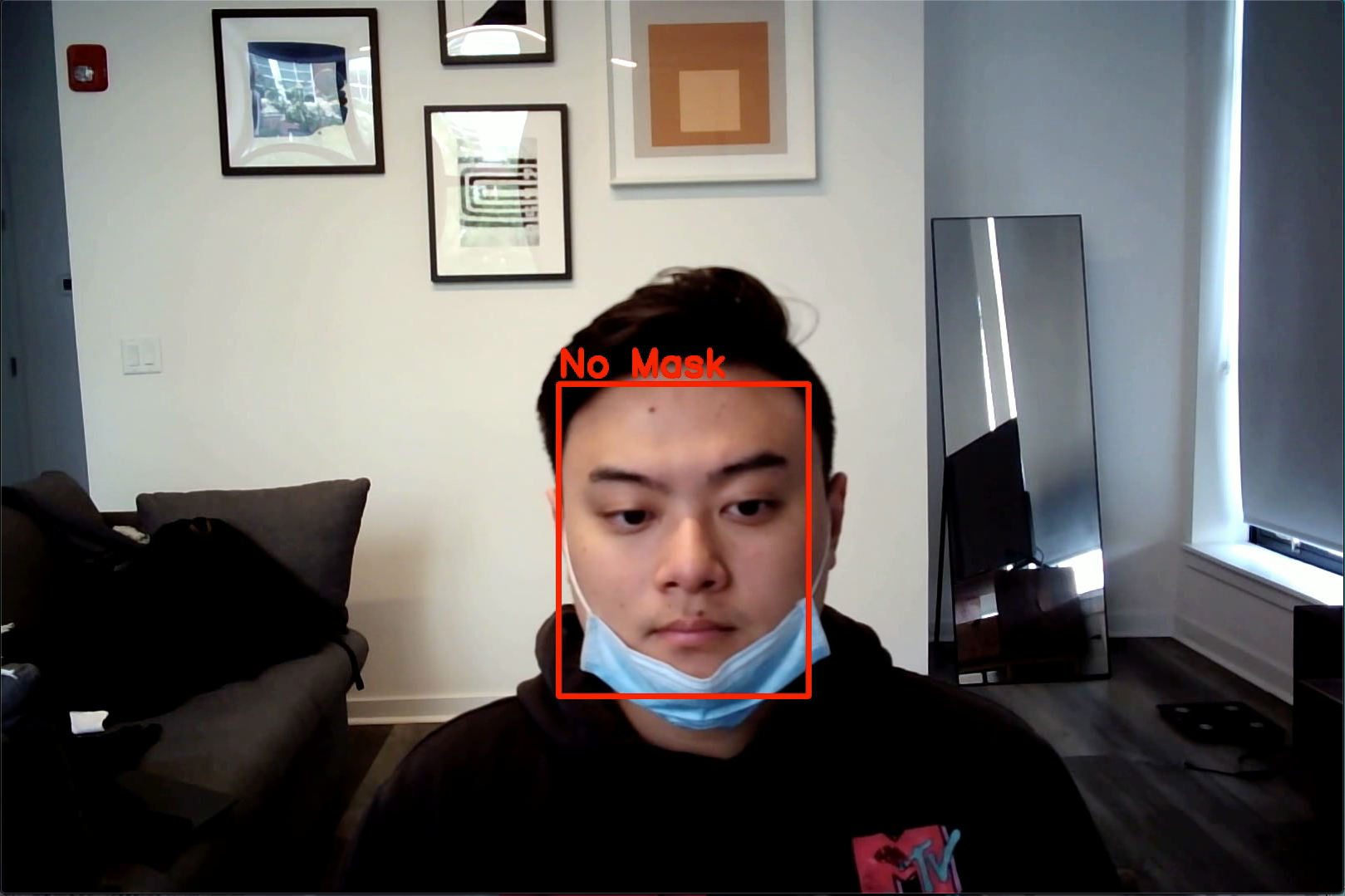}
\includegraphics[width=0.33\linewidth]{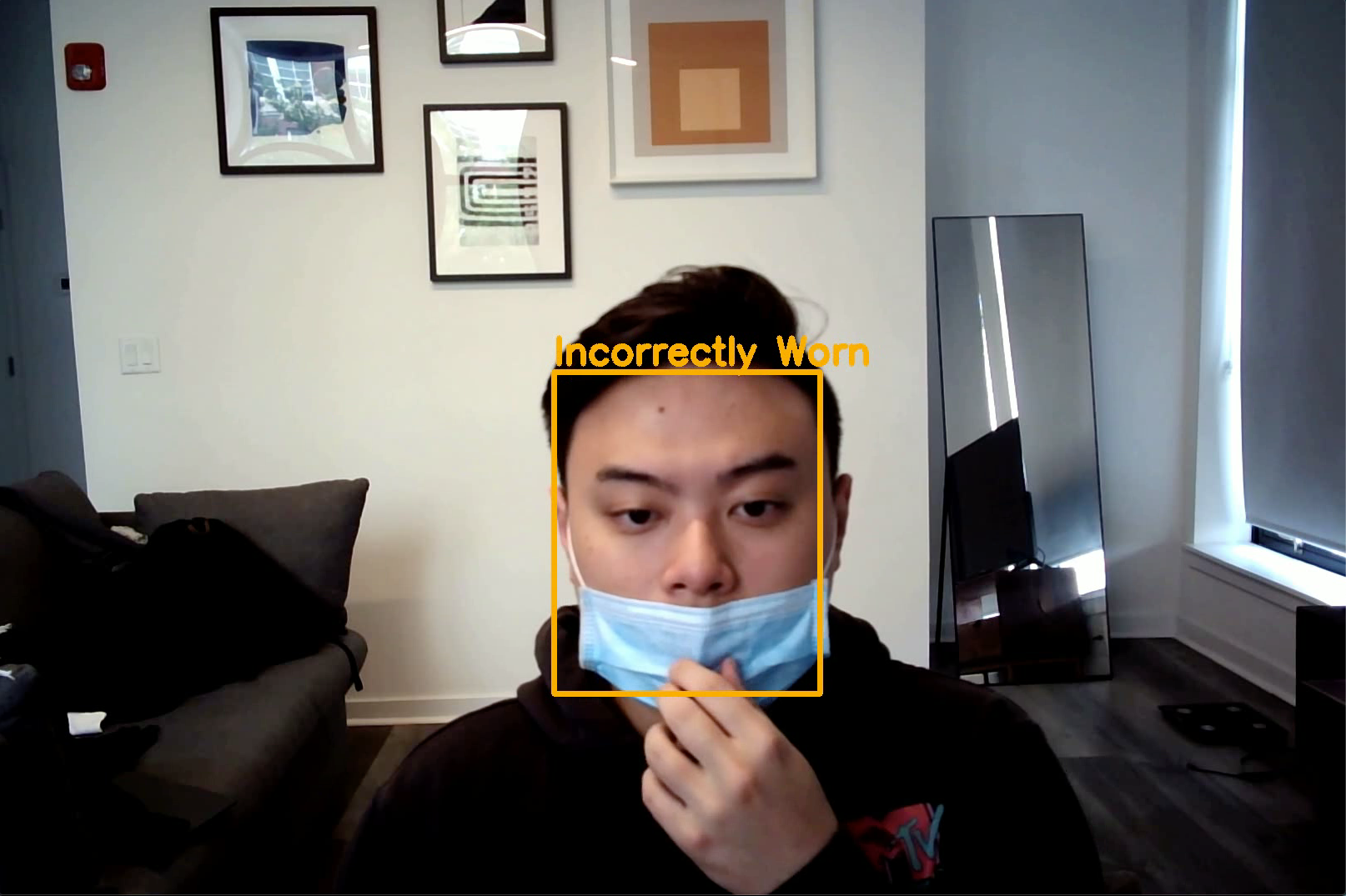}
\includegraphics[width=0.33\linewidth]{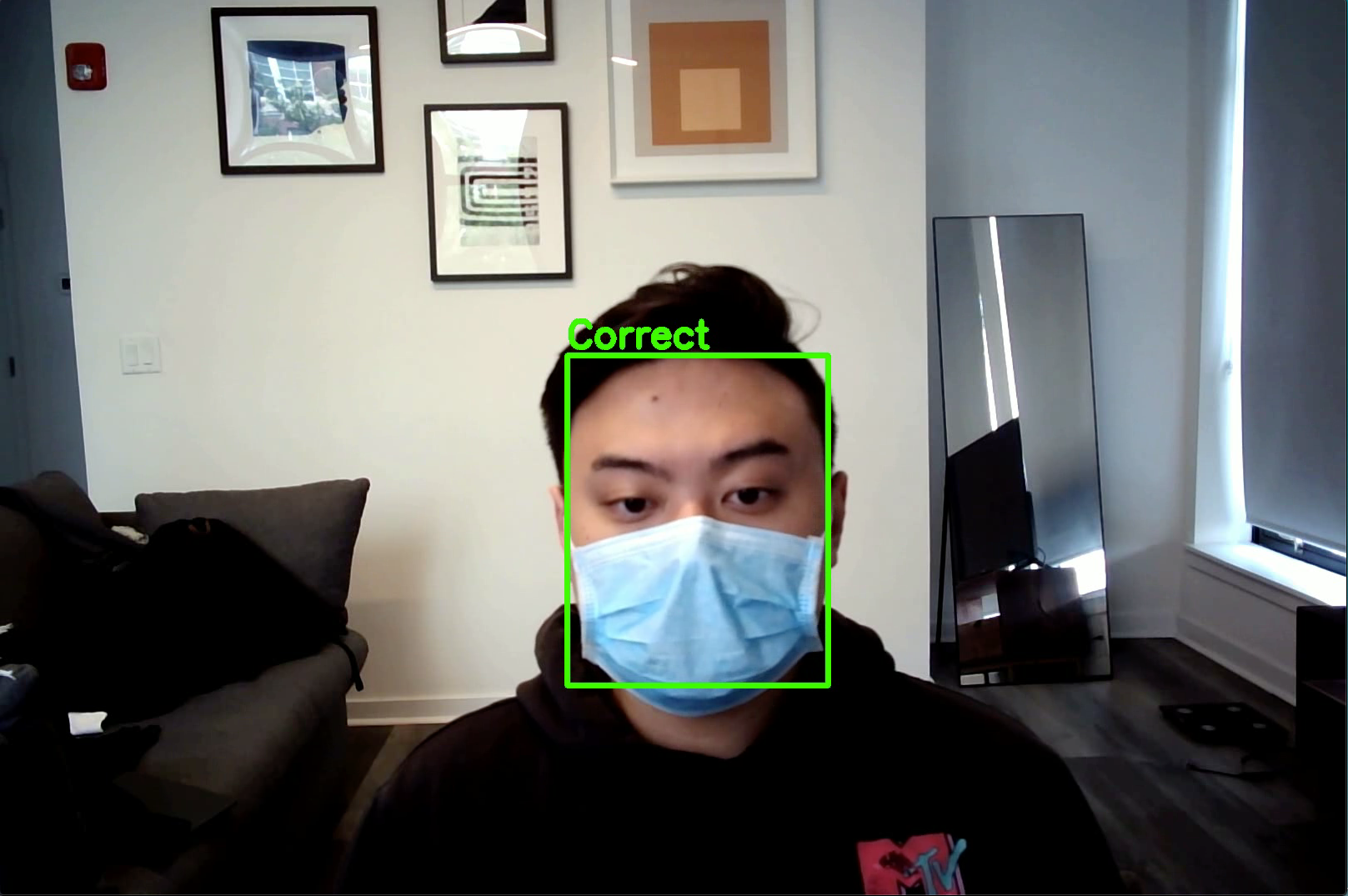}
\end{center}
   \caption{Demo Result of MTCNN + ResNet18 \cite{li_mtcnn_2021}}
\label{fig:short}
\end{figure*}
\begin{figure*}
\begin{center}
\includegraphics[width=0.33\linewidth]{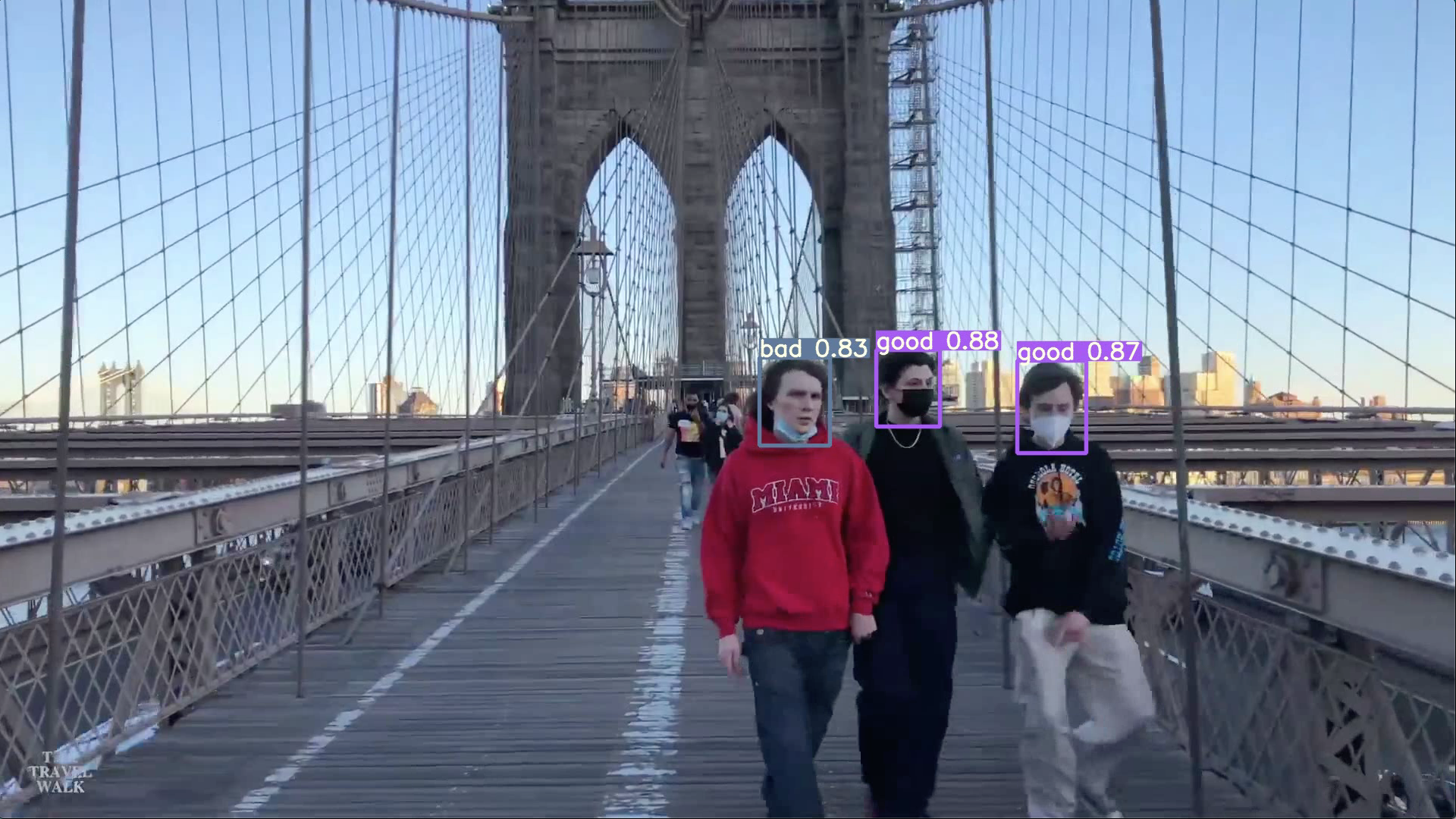}
\includegraphics[width=0.33\linewidth]{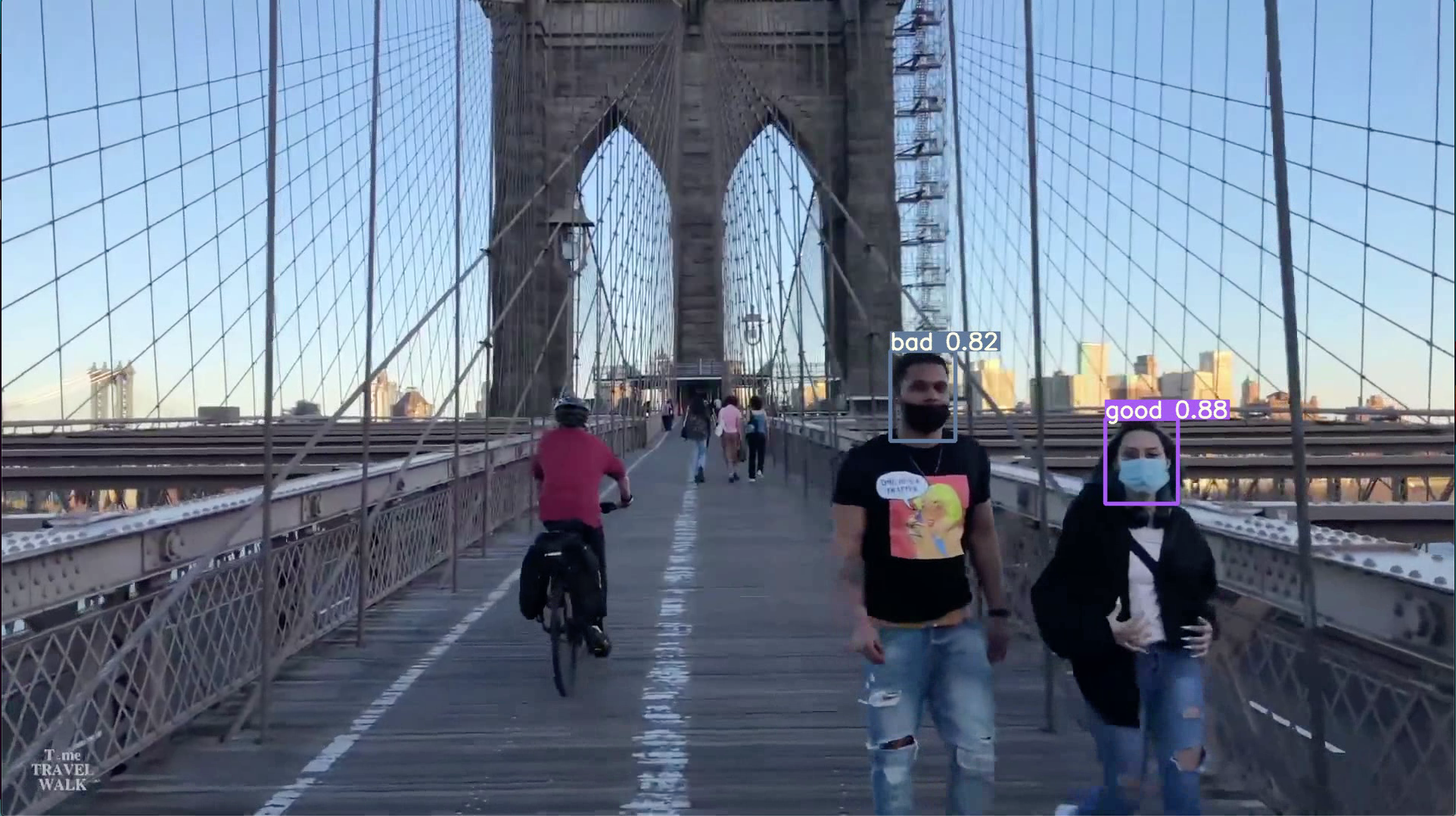}
\includegraphics[width=0.33\linewidth]{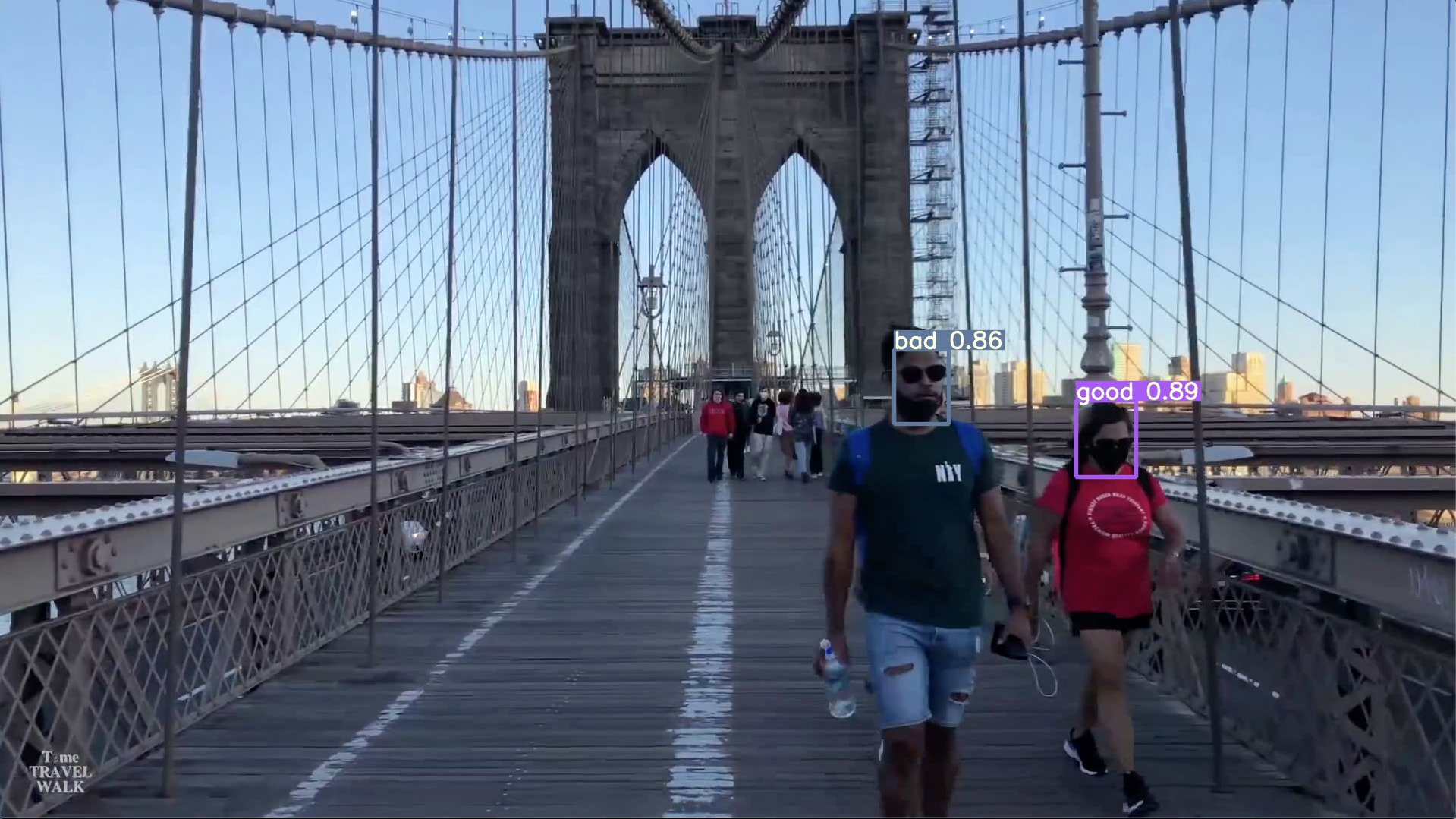}
\end{center}
   \caption{Demo Result of YOLOv5 \cite{li_yolov5_2021}}
\label{fig:short}
\end{figure*}

\section{Social Impact}
With research coming out that wearing masks would have saved countless lives \cite{peeples_face_2020} and helped stem the pandemic \cite{howard_evidence_2021}, being able to ensure compliance with mask mandates could contribute strongly to ending the pandemic around the world and preventing future ones from growing to an uncontrollable level like this one. Moreover, people may not realize they are wearing masks incorrectly, whether due to a lack of information, movement uncovering their mouth/nose, or forgetting to replace their mask when in a crowd. Tools such as these could be great for educating people on mask wearing.
Mask detection can also allow businesses and public spaces to reopen quickly and safely. This is powerful, as mental health has deteriorated throughout the pandemic \cite{Bethune_apa_2021}; by making it possible for people to live their lives and interact again, socialization will increase, which can contribute to reduced stress levels and increased resilience \cite{ozbay_social_2007}. While no argument needs to be made for why reopening the world is good, people having the choice to interact with loved ones, earn a paycheck from a rebounding economy, and participate in social activities will be a plus. If face mask detection systems can contribute to the reopening of the world and curbing of pandemics, exploring these models should be a high priority.

Of course, there are risks with such technologies. Real-time AI opens a host of security and ethical risks. If the system is not created in such a way to prevent data from being stolen, it would be possible for a hacker to steal footage, even potentially with labels showing who is or is not wearing masks, creating a privacy violation. These are already latent risks with any sort of monitoring, but they intensify with the larger system used for AI unless defenses are put in place, like using edge AI \cite{porambage_sec-edgeai_2019}. Similarly, there are safety risks in terms of malicious actors fooling the networks in order to exhaust security resources, potentially causing fatigue, avoiding detection of maskless people, or distracting from true security risks. These systems could be particular targets, given the polarization of mask wearing. We discuss these in the conclusion, as the risk profiles would need to be analyzed before deployment.

In terms of fairness, our dataset incorporates people of various backgrounds, each not, incorrectly, and correctly wearing a mask. As a result, there should be fewer issues around bias. However, it is still possible some groups are underrepresented or the algorithm has issues detecting the masks of a certain group correctly. This could unfairly treat different groups. When incorporating additional datasets or real-time data, it would be important to monitor for any inequity issues resulting.

The biggest challenge would be around authoritarianism and privacy. While this deep learning pipeline would be focused on identifying whether mask wearing is done correctly, there would likely be discussions around the trade-off of freedom versus security. There are already debates around the world in terms of pandemic restrictions and mask mandates, with many individuals wanting to decide for themselves. Tools like this one can provide governments more control over their citizens’ behavior, which would be risky. Similarly, non-government organizations, businesses, and people can already determine to some extent what sort of restrictions they want on their property, yet this automates further control that could be detrimental. For example, what if employers start firing employees who trigger multiple maskless alerts? What if those predictions were incorrect? These are real risks that could impact people’s lives. Moreover, the mask detection would need to be treated separately from any technology built on top of it, as people acclimatizing to facial mask detection could give way to organizations or governments to roll out other systems like face detection or employee monitoring. Ideally, face detection technologies would be limited to times where a pandemic is emerging or present, so that the benefits of stemming pandemics is more likely to outweigh the social risks of acclimatizing to video monitoring AI.

Given the potential benefits of saving lives, improving mental health, and keeping the world going, face mask detection technologies could have a place in ending this pandemic and preventing future ones, improving our collective well-being. At the same time, risks of bias, security, and privacy are ones that would need to be evaluated, monitored, and addressed. This is to ensure that these technologies do not give way to social ills that outweigh their benefits.

\section{Conclusion}
As observed from the performance evaluation of both pipelines, the task of face mask detection in real-time video data can definitely by automated via current deep learning models. In comparing and contrasting the two pipelines, we saw the trade-off between speed and accuracy that each model had to make. Although both pipelines are poised to gain significant boost in performance given a larger, more diverse, and better curated dataset, it is evident that single-shot object detection architectures such as YOLO is better suited for this task. Given its blazing fast inference speed, satisfactory performance on only a very small dataset, and the lack of dependency on a pre-trained face detector, the pathway to a highly-accurate and fast face mask detection system is limited only by the lack of labeled training data. This can easily be addressed by spending more money on human annotations or performing a more extensive search to combine existing data. 

As for the face detector and classifier pipeline, the two immediate areas of further work would be diversification of datasets and further distillation. The biggest gap in our training process is the dataset. We used the largest public dataset that we could find, MaskedFace-Net for correctly versus incorrectly worn masks. Those images were supplemented with Flickr Faces from NVIDIA AI for non-masked people. However, the masks were synthetically generated so they may not accurately reflect real-world settings. In addition, the ``photoshopped'' masks were all blue surgical masks, which does not reflect the diversity of masks used in the real world. As a result, finding or creating a diverse dataset of correctly, incorrectly, and non-masked people would be key for improving our model’s performance in the real world.

Our dataset was also quite homogeneous in that the photos tended to be akin to headshots. Getting a dataset of people in natural and differing environments, then labeling them, would likely improve real-world performance, as face mask detection systems would likely be deployed in varied environments beyond headshots.

Once we have a varied dataset, the next priority would be experimenting with further distillation. While our project used vanilla distillation to achieve high performance, we found that other forms of distillation are quite effective. We only had time to run distillation on a small sample of our data, but we found that KD-Lib’s implementation of noisy teacher, self learning, and messy collaboration achieved powerful results \cite{shah_kdlib_2020}. For example, running vanilla distillation, followed by noisy teacher, self learning, messy collaboration and self learning again resulted in inception’s accuracy going from 37\% to 96.5\% on a sample dataset with a couple of hundred photos. Its teacher network was ResNet with an accuracy of 67\%, which it leapfrogged. Exploring with different forms of distillation could help get the accuracy of the system closer to 100\%.

At the same time, given the importance of the performance-speed trade-off, it’s possible to try distilling progressively smaller models until it is sufficiently small and fast enough for real-world deployment. As our demonstration showed, these models can work in real-time systems already, but it is possible to reduce their overhead further.

\subsection{Long-term Implications}
Longer-term areas of further work include bias evaluation and risk analysis. Once those are addressed, deploying and monitoring in the real world would likely be the next avenue. While our dataset includes a wide representation of people across ages, genders, races, backgrounds, and other demographics, the models are not evaluated against any sort of social equity. Even with models achieving high accuracy, it is possible some groups are unfairly targeted by these models. For this reason, it would be important to do some sort of testing on how the model performs across different groups. This would likely require the images to be labeled with demographic data, and that may be a manual process. Yet, this can help avoid inequalities resulting from a real-world deployment.

In addition, prior to a real-world deployment, an assessment of risk is critical. Can these systems be misused by businesses or governments to determine private information? Can face mask detection be used in ways that are detrimental to individuals and society? What are the benefits and risks to using this technology?

Some of these go beyond the deep learning models but do concern the pipeline. For example, the security of the feeds and data is critical. Even if the model cannot do more than predict whether a mask is worn correctly, bad actors may be able to hack into the model pipeline and steal footage or even predictions to identify those masking or not. It may even be possible for someone to break in and fool the networks to set off notifications that no one is masking when that is not true. That could exhaust public resources aimed at dealing with public safety, potentially opening up risks elsewhere. The security of this pipeline will be critical.

As an extension of this idea, it could be possible for users to fool the networks using adversarial techniques. Distillation is already in our pipeline, which could help make these attacks significantly harder \cite{papernot_distillation_2016}. If these networks are not available to the public, it would be harder for attackers to be able to figure out how to fool them. However, making them commercially available and widespread would provide opportunity for attackers to access them, plus attackers would be able to test their own networks and learn to fool them. That is in addition to the other neural network attacks that exist, such as poisoning by feeding bad data. In any case, we need to be ready for the eventuality that attackers will be able to fool these networks. Various security techniques could be explored, like using generative adversarial models and pruning \cite{cheng_defending_2020}. Manually reviewing the model results periodically and applying additional approaches as needed could mitigate the risk and damage of any attacks.

Finally, with both social and technical concerns addressed, the model would be deployed to the real world. Once the deployment strategy is decided, deploying to the real world and using some sort of continuous or human-in-the-loop learning could ensure the model performs quickly to help increase correct mask wearing and curb pandemics. Finding ways to use the real-world data would also increase the size of the training data and therefore the performing models, allowing researchers and engineers to use this data to create successively better face mask detection models.

{\small
\bibliographystyle{ieee}
\bibliography{face_mask_detection}
}

\end{document}